\documentclass[letterpaper, 10 pt, conference]{ieeeconf}
\IEEEoverridecommandlockouts    %
\usepackage{graphics}           
\usepackage{times}              
\usepackage{amsmath}            
\usepackage{amssymb}            
\usepackage{graphicx}
\usepackage{algorithm}
\usepackage[noend]{algpseudocode}
\usepackage{booktabs}
\usepackage{xcolor}
\usepackage{subcaption}
\usepackage{caption}
\usepackage{siunitx}
\usepackage{todonotes}
\usepackage{hyperref}
\definecolor{instructioncolor}{rgb}{.5,.5,.5}

\usepackage[font=small]{caption}

\def\secref#1{Sec.~\ref{#1}}
\def\figref#1{Fig.~\ref{#1}}

\def\eqref#1{Eq.~(\ref{#1})}

\makeatletter
\usepackage{xspace}
\DeclareRobustCommand\onedot{\futurelet\@let@token\@onedot}
\def\@onedot{\ifx\@let@token.\else.\null\fi\xspace}

\makeatother

\usepackage{array}
\newcolumntype{L}[1]{>{\raggedright\let\newline\\\arraybackslash\hspace{0pt}}m{#1}}
\newcolumntype{C}[1]{>{\centering\let\newline\\\arraybackslash\hspace{0pt}}m{#1}}
\newcolumntype{R}[1]{>{\raggedleft\let\newline\\\arraybackslash\hspace{0pt}}m{#1}}

\newcommand{\R}[1]{\mathbb{R}^{#1}}
\newcommand{\M}{\mathtt{M}}
\newcommand{\B}{\mathtt{B}}
\newcommand{\Odo}{\mathtt{O}}

\usepackage{gensymb}

\title{\LARGE \bf Evaluation and Deployment of LiDAR-based\\ Place Recognition in Dense Forests}

\author{Haedam Oh$^{1}$ \and Nived Chebrolu$^{1}$ \and Matias Mattamala$^{1}$ \and Leonard Frei{\ss}muth$^{1,2}$ \and Maurice Fallon$^{1}$%
  \thanks{$^{1}$The authors are with the University of Oxford, UK. \texttt{\{haedam, nived, matias, mfallon\}@robots.ox.ac.uk}} 
  \thanks{Leonard Frei{\ss}muth is also with the Technical University of Munich, Germany. \texttt{\{l.freissmuth\}@tum.de}}  
  \thanks{This work has been funded by the Horizon Europe project DigiForest (101070405) and a Royal Society University Research Fellowship (M. Fallon).}%
}

\let\oldtwocolumn\twocolumn
\renewcommand\twocolumn[1][]{%
    \oldtwocolumn[{#1}{
    \begin{center}
        \includegraphics[height=7cm]{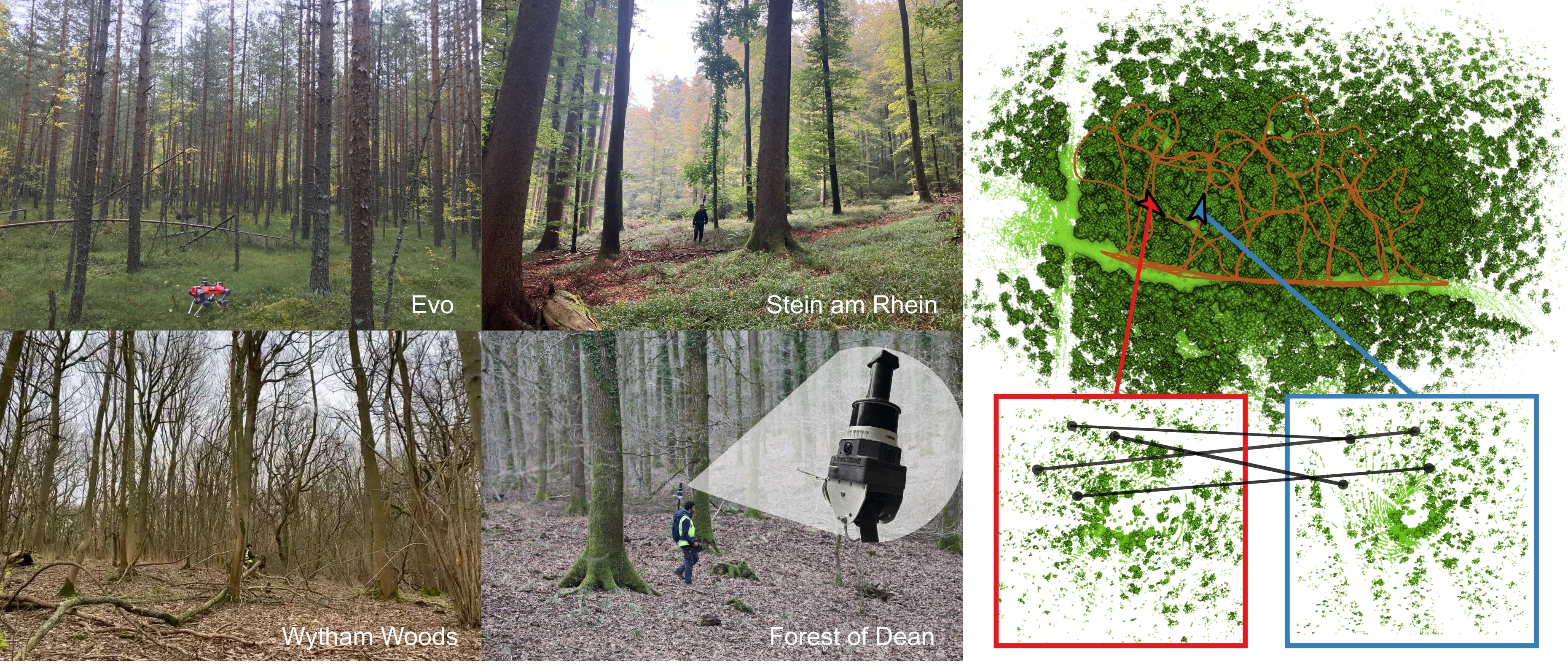}
        \captionof{figure}{In this work we evaluated LiDAR-based place recognition in dense forest environments. Left: Four different forests test sites: Evo (Finland) and Stein am Rhein (Switzerland) are coniferous forests, and Wytham Woods and Forest of Dean (UK) are deciduous. Data collection primarily utilized backpack-mounted LiDAR systems, with additional relocalization experiments conducted using a LiDAR-equipped legged robot. Right: Illustration of loop closure detection within a deeply forested area, with offset distance of \SI{11.2}{\meter} and orientation of \SI{96}{\degree} between the matched scans. The black dots indicate some corresponding locations in the two point clouds for illustration purpose.
        }
        \label{fig:motivation}
    \end{center}
}] }

\begin{document}
\maketitle
\thispagestyle{empty}
\pagestyle{empty}

\begin{abstract}
  Many LiDAR place recognition systems have been developed and tested specifically for urban driving scenarios. Their performance in natural environments such as forests and woodlands have been studied less closely.
  In this paper, we analyzed the capabilities of four different LiDAR place recognition systems, both handcrafted and learning-based methods, using LiDAR data collected with a handheld device and legged robot within dense forest environments. 
  In particular, we focused on evaluating localization where there is significant translational and orientation difference between corresponding LiDAR scan pairs. This is particularly important for forest survey systems where the sensor or robot does not follow a defined road or path. 
  Extending our analysis we then incorporated the best performing approach, Logg3dNet, into a full 6-DoF pose estimation system---introducing several verification layers for precise registration. 
  We demonstrated the performance of our methods in three operational modes: online SLAM, offline multi-mission SLAM map merging, and relocalization into a prior map. 
  We evaluated these modes using data captured in forests from three different countries, achieving \SI{80}{\percent} of correct loop closures candidates with baseline distances up to \SI{5}{\meter}, and \SI{60}{\percent} up to \SI{10}{\meter}. Video at: \url{https://youtu.be/86l-oxjwmjY} 

\end{abstract}

\section{Introduction}
\label{sec:intro}
Place recognition is an essential capability used to mitigate the effect of drift in Simultaneous Localization and Mapping (SLAM) systems, as well as to relocalize in previously mapped areas. 
LiDAR-based place recognition systems have demonstrated robustness by effectively capturing scene structure while showing low susceptibility to visual changes, making them suitable for long-term navigation. 
Previous methods have primarily focused on the problem of autonomous driving in urban environments---characterized by long linear sequences---, while their efficacy in natural settings has been much less tested.
Natural environments, specifically forests, pose different challenges: a lack of distinctive structure, loose foliage, occlusions, and limited field of view. 
This requires more robust place recognition systems to be developed than those used in urban environments. 

Recently, the Wild-Places dataset \cite{knights2023icra} has been made available to fill this gap. It consists of a large-scale forest dataset, captured by a hand-held spinning LiDAR, recording seasonal changes of the forest. The dataset was used to evaluate different state-of-the-art place recognition models in forest environments along access roads of national parks. This work showed that learning-based descriptors such as Logg3dNet \cite{vidanapathirana2022icra} showed better performance compared to handcrafted methods such as ScanContext \cite{kim2018iros}. While the findings suggested that LiDAR-based methods could provide a solution for place-based localization in forest environments, it was not conclusive if the results transferred to dense forest areas without the strong structural cues that access roads provide. 
In addition, the accuracy of the place candidates for precise 6DoF pose estimation was not assessed in forest environments, despite the opportunities that such capability offers for forestry or biodiversity monitoring applications.

In this work, we first present an evaluation of place recognition models to dense forest environments. We employ different LiDAR sensors mounted on backpack and legged robot platforms, used to collect challenging sequences enabling a rigorous analysis of existing learning-based and handcrafted LiDAR place recognition systems. Further, we present a LiDAR-based SLAM system to assess the proposed loop-closure pairs and subsequently validate them through fine-registration in both online and offline modes. Finally, we demonstrate the approach on a relocalization task, wherein the system continuously localizes its position within a prior point cloud map made up of individual LiDAR scans.

In summary, the contributions of this paper are:
\begin{itemize}

\item Evaluation of the performance of four LiDAR place recognition models, both handcrafted and learning-based, in dense forest environments.

\item Analysis of loop-closure performance as a function of the relative distance and orientation between candidate loop pairs.

\item Demonstration of the best performing method in online/single-session and offline/multi-session SLAM modes, as well as in relocalization tasks for forest inspection in a prior environment map. 

\item Release of our dense forest datasets with ground truth, collected with a backpack-mounted LiDAR across three different countries\footnote{Our datasets are available at \url{https://ori-drs.github.io/datasets/oxford-forest}}.
\end{itemize}

\section{Related Work}
\label{sec:related}

Our work is focused on LiDAR-based place recognition and localization in natural environments, particularly forests. We review some of the systems available for place recognition, which mainly involved handcrafted and learning-based systems, as well as the available datasets for their evaluation.

Among handcrafted methods, ScanContext\cite{kim2018iros,kim2021tro} stands out as a widely adopted technique that generates a descriptor by encoding the point cloud as a bird's-eye view representation. It captures height information within defined sectors and integrates them into a 2D descriptor. 
Following work extended it by incorporating additional information such as LiDAR intensity~\cite{wang2020icra} and semantics~\cite{li2021iros} to create more informative descriptors. STD~\cite{yuan2023icra} is another state-of-the-art descriptor, which encodes boundaries of planes as vertices and connects them to create multiple triangles that define the descriptor. STD operates without requiring a 360-degree scan, making it compatible with LiDAR systems with a 90-degree field of view (e.g. Livox Aria). Both ScanContext and STD can estimate a relative transformation between corresponding scans by matching their descriptors, which is useful for loop closures candidates in SLAM or relocalization tasks. However, the previous methods have been primarily tested in urban scenarios. In this work we specifically aim to study their performance in unstructured, natural environments such as forests.

Alternatively, learning-based models such as Logg3dNet~\cite{vidanapathirana2022icra}, MinkLoc3D~\cite{komorowski2021wacv}, and EgoNN~\cite{komorowski2022ral} aim to obtain distinctive descriptors from data. They mainly employ discretized representations of the LiDAR scans, and use contrastive learning schemes to learn local descriptors. This process is followed by generating a global descriptor by aggregation methods such as GeM~\cite{radenovic2019pami}, P2O~\cite{vidanapathirana2021icra}, and NetVLAD~\cite{arandjelovic2018pami}.  
Particularly, Logg3dNet~\cite{vidanapathirana2022icra} uses a sparse convolutional network with both local consistency and global scene losses learned in a contrastive manner. Similarly, EgoNN~\cite{komorowski2022ral} employs a deep CNN architecture to extract local descriptors and key points through regression, subsequently aggregating them using GeM to form a global descriptor. The local features can be used in standard feature matching and 6-DoF RANSAC registration schemes.

Alternatively, segment-based approaches such as SegMatch~\cite{dube2017icra} and SegMap~\cite{dube2019ijrr} explicitly aim to determine local segments and find  6-DoF localization based on segments. Other similar works such as NSM~\cite{tinchev2018iros} and ESM~\cite{tinchev2019ral} explicitly aimed for localization in natural environments. However, we have centered our studies on the former, single scan based place recognition-based methods which provides a similar framework to both handcrafted and learning based methods with efficient place-based candidate retrieval and a subsequent registration step for a relative transformation.

Regarding evaluation, most of the publicly available datasets for LiDAR localization aim to benchmark urban scenarios \cite{maddern2017ijrr, behley2019iccv, kim2020icra}. However, just a few datasets are centered in assessing performance in natural environments \cite{triest2022icra, knights2023icra}. In particular, the Wild-Places dataset \cite{knights2023icra} is tailored to large-scale place recognition in forests. This dataset provides point clouds and ground truth poses collected in a forest national park in Australia using hand-held spinning LiDAR at various times of the year. However, the dataset primarily focuses on specific access roads, which do not represent the dense forest areas we are interested in. We aimed to address this gap by collecting new custom datasets using a backpack LiDAR mapping device. These datasets include off-road sequences from dense forest areas across three different countries, accounting for occlusions and outliers. These provide new insights on the performance of four different place recognition models, including both handcrafted (ScanContext and STD) and learning-based (Logg3dNet and EgoNN) methods.

\section{Method}
\label{sec:main}
Our objective is to evaluate the capacity of existing LiDAR place recognition models to successfully provide robust loop closure candidates within dense forest environments. Our evaluation considers three distinct tasks: 
\begin{itemize}
  \listparindent=-20pt
  \item \emph{Task A: Online SLAM}: the proposed loop candidates contributing to a globally-consistent pose graph mapping system in an incremental manner.
  \item \emph{Task B: Offline multi-mission SLAM}: loop candidates used to link different physically overlapping missions collected at different times.
  \item \emph{Task C: Relocalization}: place recognition in a prior map made up of individual scans enabling autonomy within the map such as longer term monitoring or harvesting.
\end{itemize}

\begin{figure}[t]
  \centering
  \includegraphics[width=\columnwidth]{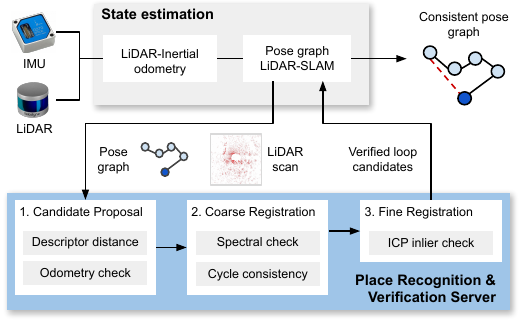}
  \caption{Our place recognition pipeline: State estimation module (VILENS) provides odometry estimates using LiDAR-inertial measurements at 10 Hz and a pose graph SLAM is used to optimize poses after successful loop closure verification. 
  The place recognition and verification server consists of three steps: 1.~loop candidate proposals, 2.~coarse registration, and 3.~fine registration. At each step, proposed loop candidates go through verification checks (grey boxes) at the global descriptor-level, local feature consistency-level, and fine registration level, respectively. A verified loop candidate is integrated in the pose graph only if it passes these three steps.}
  \label{fig:pipeline}
\end{figure}

Our system infrastructure is shown in \figref{fig:pipeline}. For state estimation, we use a LiDAR-inertial odometry system---VILENS~\cite{wisth2023tro}---, in conjunction with a pose graph SLAM framework~\cite{proudman2022ras}. Additionally, we implemented a \emph{place recognition \& verification server}, which not only provides a common interface for the different LiDAR-based place recognition models but also multi-stage verification procedures for its use in the different proposed tasks.

In the following sections, we introduce the technical details of the place recognition server, and we then present its integration to solve the three aforementioned tasks.

\subsection*{Place Recognition \& Verification Server} \label{sec:pipeline}
Our place recognition pipeline (\figref{fig:pipeline}) consists of three steps: loop candidate proposal, coarse registration, and a final fine-registration. At each step, we perform appropriate checks to filter out incorrect loop closures.

The main inputs are the pose graph, with corresponding LiDAR scans attached to each pose, as well as the single query scan. The query scan is provided by different sources depending on the task we are solving. For example, in the relocalization task it will be a live scan directly from the LiDAR sensor. Further details are provided in the corresponding sections.

\subsubsection{\textbf{Step 1: Loop candidate proposals}}
\label{subsubsec:loop-candidate}
Initial loop closure candidates are obtained by comparing global descriptors extracted from the pose graph scans as well as the query scan. In this paper, we evaluate four state-of-the-art methods for descriptor extraction: the learning-based Logg3dNet~\cite{vidanapathirana2022icra} and EgoNN~\cite{komorowski2022ral}, as well as the handcrafted ScanContext~\cite{kim2018iros} and STD~\cite{yuan2023icra}.

Given the reference pose graph and the query scan, we compute a database of descriptors using all the scans in the pose graph, given by the matrix $\mathbf{D} \in \R{N\times M}$, where $N$ is the number of poses in the pose graph and $M$ the descriptor dimension. Additionally, we compute the descriptor for the query scan, denoted by $\mathbf{d}_{q} \in \R{M \times 1}$. 

To obtain candidates, we compute the pairwise distances of the scan to the database using the cosine similarity:
\begin{equation}
  \mathbf{S} = \mathbf{D} \cdot \mathbf{d}_{q} \in \R{N \times 1}
\end{equation}
The vector of descriptor distances $\mathbf{S}$ is sorted by increasing distance, and only the top-$k$ candidates are selected using a distance threshold $\tau_{s}$, which is set by the $F_1$-max score from testing data.
If a spatial prior is available, for example from LiDAR-inertial odometry, we also perform an additional spatial check discarding all the candidates that are more than conservative estimate of \SI{20}{\meter} away from the query scan. The output is a set of candidate nodes $\{ n_c\}$.

\subsubsection{\textbf{Step 2: Coarse Registration}}
\label{subsubsec:coarse-registration}
Next, we estimate the relative 6DoF transformation that expresses the pose associated to the query scan w.r.t each candidate node, which we denote $\Delta \mathbf{T}$. For the handcrafted methods (ScanContext and STD), the relative transformation is directly an output of descriptor computation. For the learning-based approaches, we use the point-wise feature vectors outputted in the forward pass of Logg3dNet and EgoNN for feature matching, which is used in a RANSAC-based pose estimation scheme~\cite{fischler1981ransac} to estimate the relative transformation.

We additionally verify the inlier matches using the \emph{Spectral Geometric Verification}~\cite{vidanapathirana2023ral} method, which provides an additional measure of the quality of the feature matches.

\begin{figure}[t]
  \centering
  \includegraphics*[width=\columnwidth]{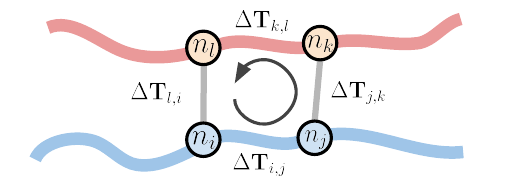}
  \caption{Our proposed cycle consistency check is general and applies to the online and offline multi-mission SLAM case, as well as relocalization tasks. We only need the relative transformation estimates and loop candidates between four nodes $n_i, n_j, n_k, n_l$ to verify the validity of a loop. Please refer to \secref{subsubsec:coarse-registration} for technical details.}
  \label{fig:cycle-consistency}
\end{figure}

Lastly, we carry out a \emph{cycle consistency} verification (\figref{fig:cycle-consistency}), which checks whether the relative transformations between pairs of nodes are mutually consistent with one another. Given four pose graph nodes $n_i, n_j, n_k, n_l$, we test how close the following equivalence holds:
\begin{equation}
\Delta\mathbf{T}_{i,j}\, \Delta\mathbf{T}_{j,k}\, \Delta\mathbf{T}_{k,l}\, \Delta\mathbf{T}_{l,i}\, \approx \mathbf{I}_{4\times4} 
\end{equation}
If this difference is more than a threshold of \SI{10}{\centi\meter} or \SI{1}{\degree} we reject the candidate. This is shown in \figref{fig:cycle-consistency}, please refer to the corresponding sections for further details.

The interpretation of these transformations change if we discuss online SLAM (\secref{sec:online_slam_mode}), offline multi-mission SLAM (\secref{sec:offline}), or pure relocalization (\secref{sec:relocalization}). 

\subsubsection{\textbf{Step 3: Fine Registration}}
\label{subsubsec:fine-registration}
Finally, we employ the Iterative Closest Point (ICP) algorithm~\cite{besl1992icp} for fine registration of the proposed candidates. We use the \emph{libpointmatcher} implementation~\cite{pomerleau2013iros}, which also provides information on the quality of the registration, such as the proportion inlier points and the residual error of each point to access the alignment quality.
We use a residual error of \SI{20}{\centi\meter} as a threshold which we observed to be effective over different forest environments.
The verified candidates are then used for SLAM or relocalization tasks, which are detailed in the following sections. %

\subsection*{Task A: Online Single-mission SLAM} 
\label{sec:online_slam_mode}
\begin{figure}[t]
  \centering
  \includegraphics[width=0.99\columnwidth]{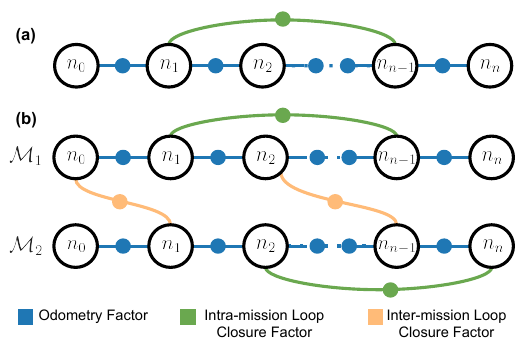}
  \caption{Pose graph formulation used for (a) online, and (b) offline multi-mission SLAM optimization. Each node $n_{i}$ has a 6DOF pose $\mathbf{x}_{i}$, which correspond to the main variables estimated on each case.}
  \label{fig:factor_graph}
\end{figure}

The first task we consider is LiDAR-based online SLAM. Our implementation defines it as an incremental pose graph estimation problem (see \figref{fig:factor_graph},(a)). Consider consecutive loop closures at nodes $n_{i}$, $n_{i+1}$ and $n_{j}$, $n_{j+1}$. Edges are provided by relative estimates from our LiDAR-inertial odometry system (odometry factors, denoted by $\Delta\mathbf{T}_{i,i+1}, \Delta\mathbf{T}_{j, j+1}$), and verified loop closure candidates from our place recognition server (loop closure factors, $\Delta\mathbf{T}_{i+1, j}, \Delta\mathbf{T}_{i, j+1}$).

For the cycle consistency verification described in \secref{subsubsec:coarse-registration}, we consider the relative transformation change between consecutive loop closure candidates w.r.t the pose graph poses and the odometry change (\figref{fig:cycle-consistency}  $i$, $j$, $k$, $l$, replaced by ${i}$, ${i+1}$, ${j}$, ${j+1}$). Again, a cycle consistency needs to be satisfied:
\begin{equation}
  \label{eq:cycle-online}
  \Delta\mathbf{T}_{i,i+1}\, \Delta\mathbf{T}_{i+1, j}\, \Delta\mathbf{T}_{j, j+1}\, \Delta\mathbf{T}_{i, j+1}^{-1} \approx \mathbf{I}_{4\times4}
\end{equation}
\subsection*{Task B: Offline Multi-Mission SLAM}\label{sec:offline}
Offline multi-mission SLAM addresses the challenge of merging multiple pose graph SLAM missions ${\mathcal{M}_{1, \ldots, n}}$, collected over time, with partly overlapping areas. 
The goal is to find inter-mission loop candidates to construct a unified map in a common reference frame. This application is relevant for forestry applications, where it is required to map larger areas by integrating surveys conducted over multiple missions or campaigns.

Unlike the scenario of on-road navigation, where similar routes (hence locations) are revisited, we consider off-road scenarios where the missions are collected in dense forests, where it is often unfeasible to retrace the same paths on each sequence. To avoid inefficiently retracing our steps, we wish to identify loop candidates when passing no closer than about \SI{10}{\meter}, providing the flexibility needed to merge two roughly overlapping missions.

Each mission ${\mathcal{M}_{i}}$ is defined by a pose graph with odometry factors and intra-mission loop closures, obtained during each independent online SLAM run. We aim to provide additional \emph{inter-mission} loop candidates that bridge nodes across missions, as shown in \figref{fig:factor_graph} (b). In this case, potential \emph{inter-mission} loop candidates are obtained through successive one-on-one matching of each mission's nodes against one of other missions, integrating them into a unified pose graph.
For the loop proposal step, we execute the same procedures described in \secref{subsubsec:loop-candidate}.
The cycle consistency step considers pairs of nodes within the same mission, namely $n_i, n_j \in \mathcal{M}_1$ and $n_k, n_l \in \mathcal{M}_2$. The intra-mission relative transformation are then $\Delta\mathbf{T}_{i,j}, \Delta\mathbf{T}_{k, l}$, while the inter-mission relative transformations between loop candidates are given by $\Delta\mathbf{T}_{i,k}, \Delta\mathbf{T}_{j,l}$:
\begin{equation}
  \label{eq:cycle-offline}
  \Delta\mathbf{T}_{i,j}\, \Delta\mathbf{T}_{j,l}\, \Delta\mathbf{T}_{k, l}^{-1}\, \Delta\mathbf{T}_{i,k}^{-1}\, \approx \mathbf{I}_{4\times4}
\end{equation}

\subsection*{Task C: Relocalization} \label{sec:relocalization}
Lastly, we consider the case in which a prior map of the forest is available, e.g. from online SLAM. Our place recognition \& verification server can then be used as a relocalization module, by using the loop candidate proposals to produce initial pose estimates and then executing coarse-to-fine registration. This enables real-time localization of the LiDAR sensor's base $\B$, with the prior map's coordinate frame $\M$, denoted by $\mathbf{T}_{\M\B}$.

Similarly to the previous tasks, the main difference is in defining the cycle consistency check. For this case, it is between the current and the last successful relocalization: Given the last relocalization estimate $\mathbf{T}_{\M\B}(t-1)$ and the current estimate $\mathbf{T}_{\M\B}(t)$, we compared them against the odometry estimates at the same timestamps $\mathbf{T}_{\Odo\B}(t-1)$ and $\mathbf{T}_{\Odo\B}(t)$, where $\Odo$ indicates the fixed odometry frame. The cycle consistency check is then defined as:

\begin{equation}
  \label{eq:cycle-relocalization}
  \underbrace{\mathbf{T}_{\M\B}(t)^{-1}\, \mathbf{T}_{\M\B}(t-1)}_{\Delta\mathbf{T} \text{ in $\M$ frame}}  \, \underbrace{ \mathbf{T}_{\Odo\B}(t-1)^{-1}\,  \mathbf{T}_{\Odo\B}(t)}_{\Delta\mathbf{T} \text{ in $\Odo$ frame}} \approx \mathbf{I}_{4\times4}
\end{equation}
This check is used to verify the relocalization estimate, and if successful, ICP is used to fine-localize the LiDAR sensor only against the corresponding individual map scan (instead of the full map point cloud).

This relocalization capability could facilitate various applications, such as enabling a harvester robot to operate autonomously within a prior map or enabling foresters to visualize a rendering of the virtual forest along with important information on a screen in real-time. An example demonstrating the latter is presented in \secref{sec:exp_relocalization}, where a prior map of the forest is generated using a backpack-based LiDAR mapping system, and a legged robot continuously relocalizes itself within that prior map as part of a teleoperated inspection task.

\section{Experimental Evaluation}\label{sec:exp}
Our evaluation consisted of four test sites featuring different forest compositions. 
We used two different lidar models: a Hesai XT32 LiDAR---\SI{50}{\meter} effective range and \SI{30}{\degree} narrow field of view and a Hesai QT64---\SI{30}{\meter} range, \SI{100}{\degree} wide field of view---mounted on a backpack. Forest environments are visualized in \figref{fig:motivation}. 
\begin{itemize}
  \item Evo (Finland), Hesai XT-32, characterized by tall, a mix of broad-leaf and coniferous trees. 
  \item Stein am Rhein (Switzerland), Hesai XT-32, thinned coniferous trees, open canopy, and flat ground. 
  \item Wytham Woods (UK), Hesai QT-64, a very dense forest with cluttered trees with mixed species and ground vegetation, as well as uneven terrain (valleys and hills). 
  \item Forest of Dean (UK), Hesai QT-64, less dense broad-leaf, oak trees and flat grounds.
\end{itemize}

We evaluated all three operational tasks of our system: online SLAM, offline multi-mission SLAM, and relocalization. While the online SLAM task demonstrated the capability of detecting large baseline loop closures reliably, the offline multi-mission SLAM task and relocalization task demonstrate the application of our system to large-scale mapping and forest inventory surveying.

\begin{figure}[t]
  \centering
  \includegraphics[width=0.99\linewidth]{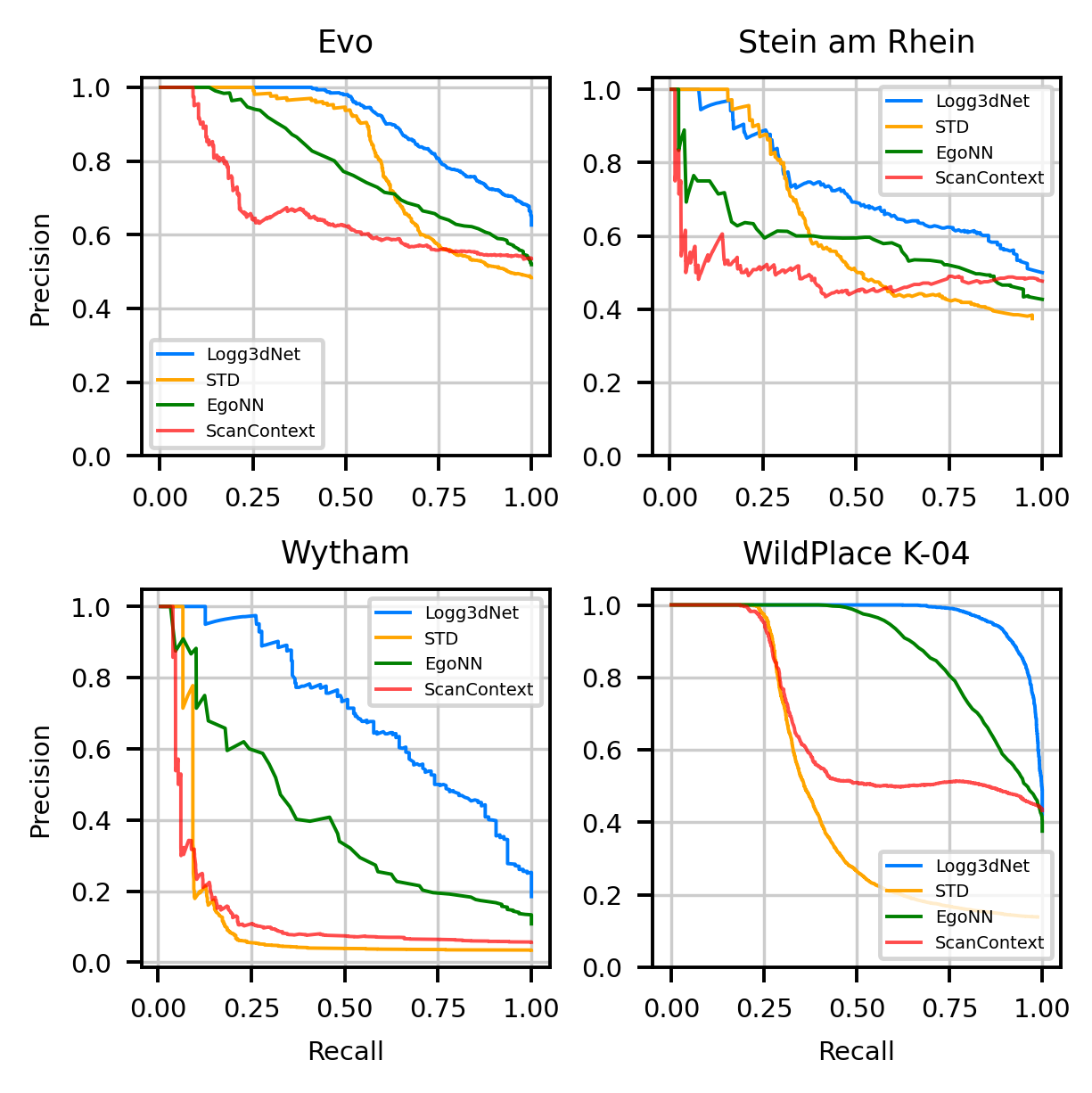}
  \caption{Precision-recall curves obtained for Logg3dNet, STD, EgoNN and Scan Context in our four dense forest datasets. Only the top-1 candidate within \SI{10}{\meter} of the ground truth position is regarded as a true positive candidate.}
  \label{fig:pr_curves}
\end{figure}

\begin{figure}[t]
  \centering
  \includegraphics[width=0.90\linewidth]{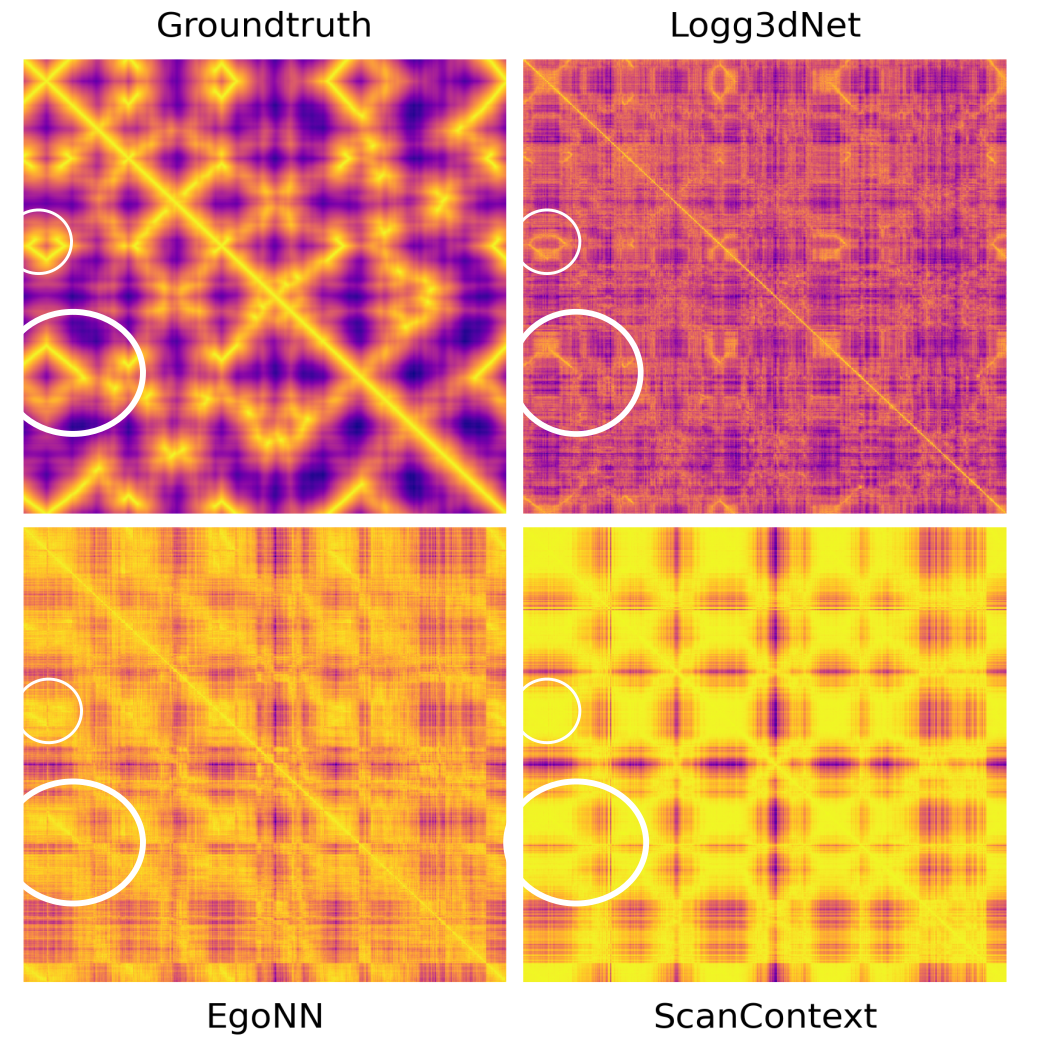}
  \caption{Heatmaps depicting descriptor distances for the Evo dataset. Yellow hues denote a high descriptor similarity between scans, whereas purple indicates low similarity. Patterns more closely resembling the ground-truth (top-left) indicate better descriptor performance. Logg3dNet descriptors show the most similar patterns, whereas ScanContext descriptors are the least discriminative among these models.}
  \label{fig:heatmap_evo12}
\end{figure}

\subsection{Place Recognition Descriptors}
\label{sec:exp_desc_analysis} 
In this experiment, we evaluated the descriptors of four different place recognition models (Logg3dNet, EgoNN, ScanContext, STD) focusing on their ability to accurately capture loop-candidates in forest environments. We used poses obtained from our SLAM system as the ground truth. Logg3dNet and EgoNN models are learning-based methods and were pre-trained on the Wild-Places dataset. 

We measured precision-recall curves to assess the capability of detecting correct loop candidates at various descriptor distance thresholds $\tau_{s}$ in four different forests. We consider loop closures within a distance threshold of \SI{10}{\meter} to be correct.  

From our obtained precision-recall curves \figref{fig:pr_curves}, it is evident that Logg3dNet consistently outperforms the other models across the four different forests. Particularly, on the Evo and Stein am Rhein datasets, Logg3dNet showed the best performance both in terms of precision and recall, without experiencing any drastic drops in precision. In contrast, ScanContext demonstrated a significant decrease in precision, attributed to its dependency on the vertical field of view of the input scan. 

In more challenging scenarios, such as Wytham Woods, handcrafted models showed a notable decline in performance. However, Logg3dNet remained at the top, successfully retrieving a substantial portion of correct loop-candidates, achieving a 70\% precision at a 50\% recall rate.

To further analyze the distinctiveness of each descriptor, we measured the descriptor distances between all query and database descriptors. This is shown in \figref{fig:heatmap_evo12} as a heatmap, which provides a visual representation of the discriminative potential of each descriptor. Consistent with the precision-recall curves, Logg3dNet descriptors exhibited higher similarity with the ground truth heatmap as observed in the highlighted areas, indicating a high true-positive rate and low false-positive rate, respectively. This implies that Logg3dNet descriptors can effectively detect corresponding loop-candidates during revisits, whereas EgoNN and ScanContext tend to be less discriminative, often returning numerous false-positive candidates. Based on this evidence, we chose Logg3dNet as main the place recognition method for the rest of the experiments.

\subsection{Online Place Recognition}
In this experiment, we investigated the online place recognition capability of our system, wherein loop closures from the place recognition module are integrated into the SLAM system. The database $\mathbf{D}$, is incrementally built as the sensor moves through the environment. When matching, we exclude the most recent 30 seconds of data to prevent loop closures with immediately recent measurements.

\begin{figure}[t]
  \centering
  \includegraphics[width=0.99\columnwidth]{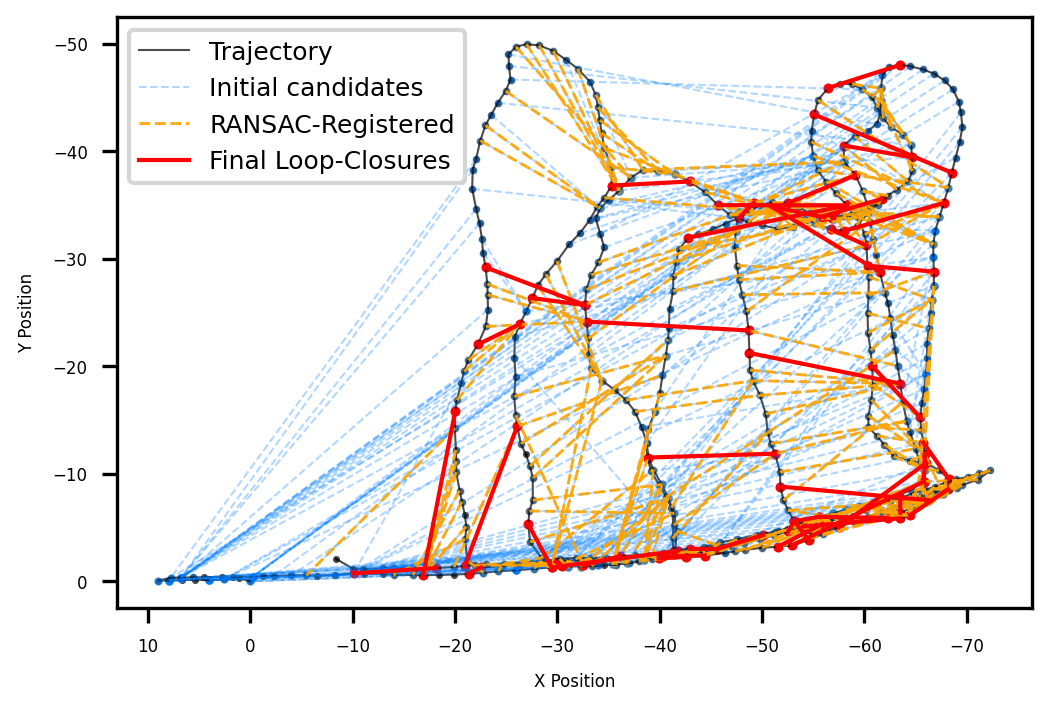}
  \caption{Online place recognition on a sequence of the Evo dataset. The sequence covers about 25 minutes of walk with the backpack mapping system (Hesai-XT32). Bold red lines show the loop closures integrated into the SLAM system, successfully identified up to distance of \SI{17}{\meter} within dense forest areas.}
  \label{fig:exp_2_1_online_place_recognition}
\end{figure}

\figref{fig:exp_2_1_online_place_recognition} presents an illustrative example of online SLAM performance on the Evo dataset, depicting the sets of loop candidates after each verification step. Initially, many loop closure candidates are proposed (shown in blue) under a descriptor matching threshold $\tau_{s}$ of $F_1$-max score. Loop closures beyond a conservative estimate of \SI{20}{\meter} are rejected using the odometry information. After this, a subset of loop closure candidates are identified using RANSAC matching (highlighted in orange), and finally, a refined set of loop closures that pass the consistency and ICP steps are integrated into the SLAM framework. Final loop closures (shown in red) are one of ICP verified loop candidates by checking pose graph density to avoid over-constraining the pose graph.

\begin{figure}[t]
  \centering
  \includegraphics[width=0.99\columnwidth]{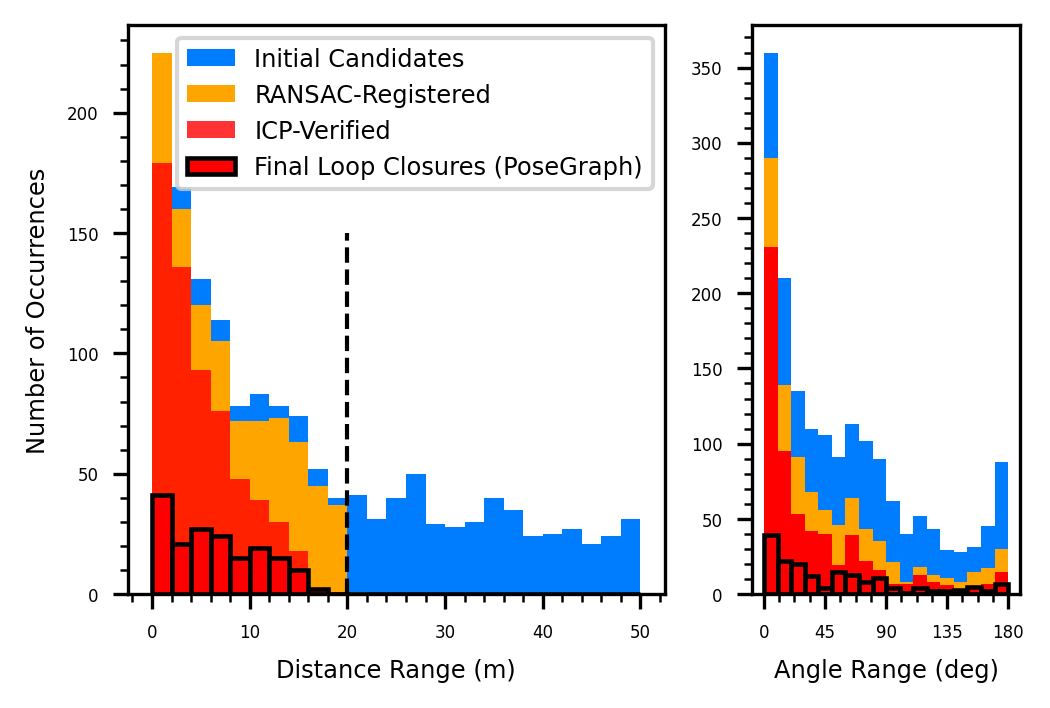}
  \caption{Loop closures distribution by distance and angle at various stages of the pipeline on Evo dataset.
   Initial candidates based on descriptor distance are shown in blue. Candidates beyond \SI{20}{\meter} are rejected using odometry information. Candidates within \SI{20}{\meter} undergo RANSAC pre-registration with additional verification steps of SGV\cite{vidanapathirana2023ral} and pairwise checks (Yellow). Then these candidates are refined using ICP fine-registration (Red), and final loop closures in the pose graph after checking constraints density in pose graph (red with black outlined).}
  \label{fig:exp_2_2_loop_closure_histograms}
\end{figure}

Next, we conducted a comprehensive analysis of loop closure statistics based on distance and viewpoint angles, shown \figref{fig:exp_2_2_loop_closure_histograms}. Our findings show that the system can successfully identify loop closure pairs across considerable baseline distances (\SIrange{10}{20}{\meter}). We observed that despite the large baseline, a significant portion of initial candidates can be registered using RANSAC-based matching, indicating that the correspondences are accurate. However, the proportion of candidates verified by ICP decreases as the distance between scans increases. Specifically, when scans are \SI{10}{\meter} apart, $\sim$\SI{60}{\percent} of RANSAC-registered candidates are successfully verified by ICP, and when \SI{15}{\meter} apart, only $\sim$\SI{40}{\percent} remains verified. This decrease is due to the diminishing overlap ratio between corresponding scans with increasing distance, making convergence of ICP challenging.

Similarly, in terms of viewpoint orientation difference, we observed that about $\sim$\SI{60}{\percent} of loop candidates up to \SI{90}{\degree} difference are verified both at the RANSAC-registration and ICP-based checks. However, we observed a degradation of performance over \SI{90}{\degree}, which can be attributed to the occlusions in the scans present at large orientation differences. Nonetheless, despite this degradation, the number of final loop closures integrated into the SLAM system proved to be sufficient for correcting the drift.

\begin{figure*}[t]
  \centering
  \includegraphics[width=0.99\linewidth]{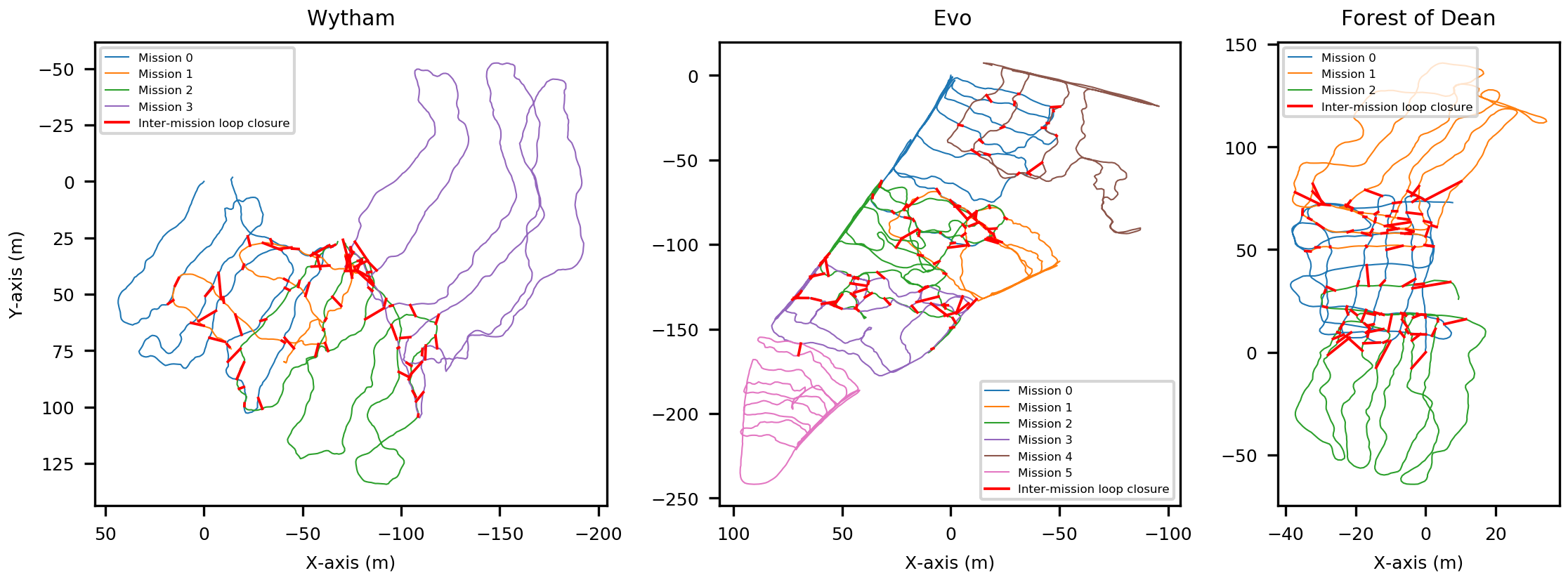}
  \caption{Offline multi-mission SLAM. Left: Wytham - a densely wooded area with uneven terrain, including hills. Center: Evo - featuring a LiDAR setup on an incline, with loop closures occurring primarily when viewpoints are closely aligned. Right: Forest of Dean - flatter terrain compared to Wytham, with a sparser plantation, allowing more frequent loop closures within intersecting areas.}
  \label{fig:exp_multi_mission}
\end{figure*}
\subsection{Offline Multi-Mission SLAM} 
\label{exp:offline_multi_mission}
This experiment showcases the ability of our approach to obtain loop closures between different mapping missions and to merge those missions into a common map. 

Figure \ref{fig:exp_multi_mission} presents the results of merging different sequences within three different datasets: Wytham Woods, Evo, and the Forest of Dean. Each individual mission covers approximately one hectare, with merged map areas ranging from three to five hectares.
We observed that there were more frequent \emph{intermission} loop closures in the Forest of Dean compared to Wytham in overlapping areas. This can be attributed to the higher tree density, foliage, and vegetation present in Wytham, making the descriptors less distinctive.     

Further, we tested the robustness of our system on the Evo multi-mission dataset, where the XT32 LiDAR was inclined aimed at capturing the forest canopy. Despite the asymmetry in point clouds introduced by this inclination change, which primarily captured points in the forward direction and missed points from the back of the LiDAR, our approach successfully identified loop closures and achieved multi-mission map merging.

Overall, our experiments showed potential for efficient large-scale mapping, as we did not require to start at the open access roads nor following exactly the same paths to achieve loop closures between inter-missions. We also built the map incrementally, one section at a time, hence overlapping areas could only be guaranteed for subsequent missions.

\begin{figure}[t]
  \centering
  \includegraphics[width=0.90\linewidth]{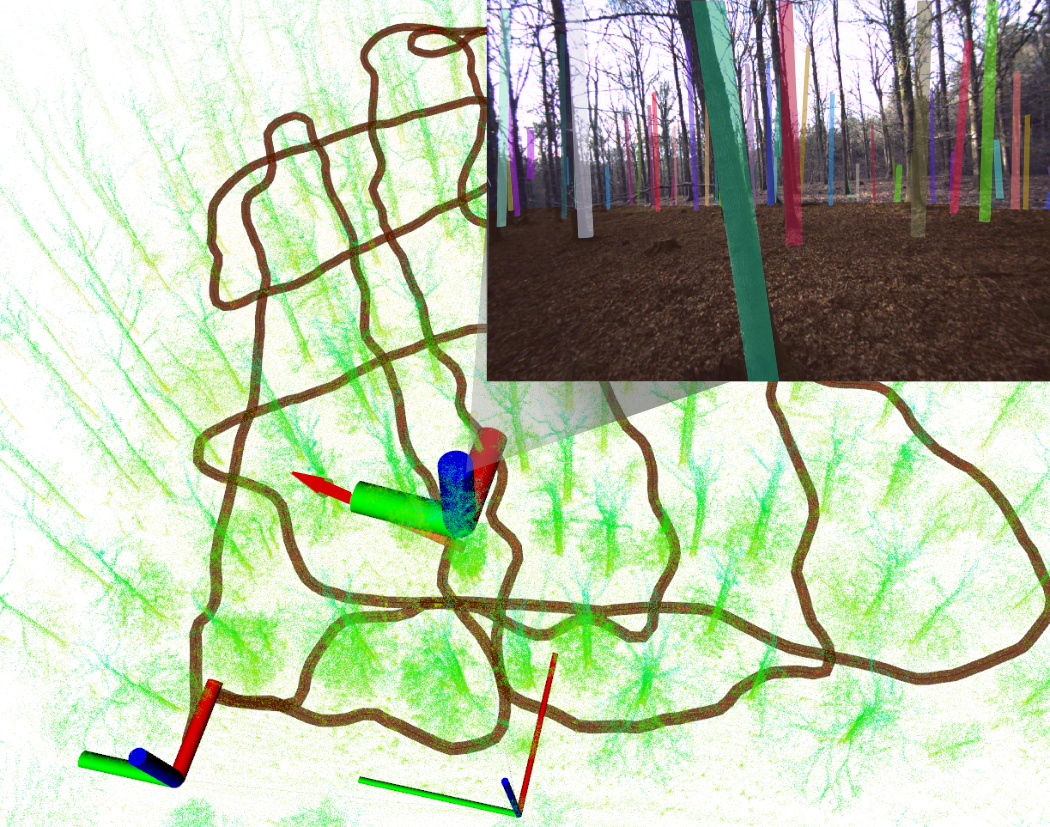}
  \caption{Demonstration of relocalization capability. The LiDAR sensor, illustrated by the thick frames, is relocalized in a prior map. We rendered a virtual view of the forest digital map synchronized with images from our camera (right). Please refer to our supplementary video for a demonstration.
  }
  \label{fig:relocalization}
\end{figure}

\subsection{Relocalization} 
\label{sec:exp_relocalization}
This experiment showcases our relocalization capability in a dense forest. In this scenario, we successfully localize without any prior information, and continuously track the pose of the LiDAR sensor within the pose graph-based SLAM map (avoiding requiring a single monolithic point cloud).
Figure \ref{fig:relocalization} presents an illustrative example of this capability using a sequence of the Forest of Dean dataset. The previous map was built using a backpack-LiDAR sequence, while the relocalization sequence was obtained with a LiDAR mounted on a teleoperated legged robot. To further demonstrate this capability, we rendered a simplified virtual forest (again obtained from the backpack prior map). In our supplementary video we show how we can exploit the accurate localization estimate from our system to generate a real-time rendering of the forest and Diameter at Breast Height (DBH) as the device moves. This application demonstrates the potential of our localization system to support innovate forestry tasks such as forest inventory or virtual reality.

\subsection{Study of ICP inlier-based check}
\label{sec:exp_icp_ablation}

\begin{figure}[t]
  \centering
  \includegraphics[width=0.99\linewidth]{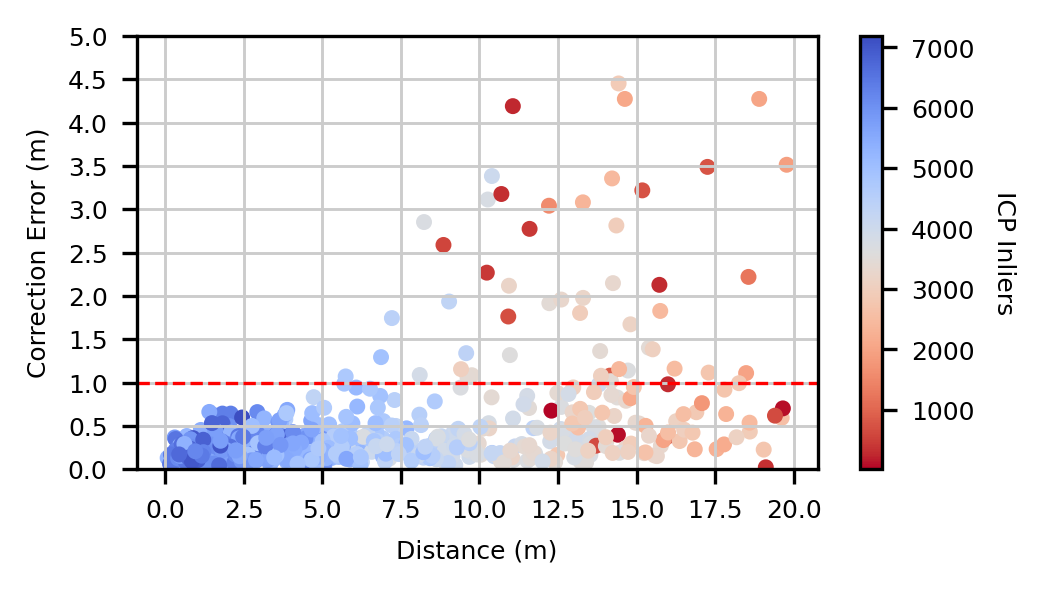}
  \caption{Analysis of final ICP registration check. X-axis shows the distribution of loop-candidates by distance after ICP.
  Y-axis shows ICP correction error w.r.t. the coarse-registration from RANSAC. Color indicates the number of ICP inliers with 30 iterations and RMSE=\SI{0.01}{\meter} out of 20k points. Above \SI{10}{\meter}, the inliers numbers decreased below 4k points, and the correction error is no longer bounded to \SI{1}{\meter}. We proposed to set a threshold on the correction distance to ensure that only high-quality candidates are integrated in the pose graph optimization.}
  \label{fig:ablation_icp_inliers}
\end{figure}

Our final experiment investigated the ICP inlier-based check, used to determine the loop closure candidates that are integrated in the pose graph optimization.
\figref{fig:ablation_icp_inliers} illustrates the corrections---applied on top of the initial transformation prior---as estimated by the ICP registration at different distances. The color encodes the number of inlier points obtained during the registration process, with blue indicating a large number of inliers and red indicating a smaller number. As expected, the candidates with more inliers corresponded to the ones that are also closer to the reference poses (smaller distances). In contrast, loop closures occurring beyond \SI{10}{\meter} that propose a substantial transformation correction often have fewer inlier ICP points and are thus less reliable. Based on this analysis, we established an inlier threshold on the proposed correction distance---we used a value of \SI{1}{\meter} (dotted red line). We observed that this threshold helped reduce the number of incorrect loop closures from being integrated into the pose graph.

\section{Conclusion}
\label{sec:conclusion}

In this paper, we conducted extensive testing of LiDAR place recognition systems in dense forest environments. We presented a place recognition and verification system that leverages three stages of loop-candidate verification: at the global descriptor-level, local feature-level, and fine point cloud level. These place recognition modules were seamlessly integrated into a pose graph SLAM system and evaluated across three distinct scenarios: online SLAM, offline multi-mission SLAM, and relocalization. Our experiments provide further insights on the performance of currently available LiDAR-based place recognition methods in dense forests. Further, they demonstrate different integration cases to achieve 6-DoF localization, opening future applications for forest inventory, inspection, and autonomous tasks.

\section*{Acknowledgments}
We thank the assistance of the Swiss Federal Institute for Forest, Snow and Landscape Research (WSL), Forestry Research UK, and PreFor for access to forest sites. 

\bibliographystyle{plain_abbrv}

\bibliography{new}

\end{document}